\documentclass[journal]{vgtc}                     

\onlineid{1090}

\vgtccategory{Research}

\vgtcpapertype{Original Article}

\title{HEADSET: Human Emotion Awareness under Partial Occlusions Multimodal DataSET}

\author{%
  \authororcid{Fatemeh Ghorbani Lohesara}{0000-0002-5121-3052},
  Davi Rabbouni Freitas, Christine Guillemot, Karen Eguiazarian, and 
  Sebastian Knorr
}

\authorfooter{
  \item
  	Fatemeh Ghorbani Lohesara is with Communication Systems Group at Technische Universität Berlin.
  	E-mail: ghorbani.lohesara@tu-berlin.de.
  \item
  	Davi Rabbouni Freitas is with INRIA.
  	E-mail: davi-rabbouni.de-carvalho-freitas@inria.fr.

  \item Christine Guillemot is with INRIA.
  	E-mail: christine.guillemot@inria.fr.

   \item
  	Karen Eguiazarian is with Computational Imaging Group at Tampere University.
  	E-mail: karen.eguiazarian@tuni.fi.

  \item Sebastian Knorr is with Ernst-Abbe University of Applied Sciences Jena.
  	E-mail: sebastian.knorr@eah-jena.de.
}

\abstract{%
The volumetric representation of human interactions is one of the fundamental domains in the development of immersive media productions and telecommunication applications. Particularly in the context of the rapid advancement of Extended Reality (XR) applications, this volumetric data has proven to be an essential technology for future XR elaboration. In this work, we present a new multimodal database to help advance the development of immersive technologies. Our proposed database provides ethically compliant and diverse volumetric data, in particular 27 participants displaying posed facial expressions and subtle body movements while speaking, plus 11 participants wearing head-mounted displays (HMDs). The recording system consists of a volumetric capture (VoCap) studio, including 31 synchronized modules with 62 RGB cameras and 31 depth cameras. In addition to textured meshes, point clouds, and multi-view RGB-D data, we use one Lytro Illum camera for providing light field (LF) data simultaneously. Finally, we also provide an evaluation of our dataset employment with regard to the tasks of facial expression classification, HMDs removal, and point cloud reconstruction. The dataset can be helpful in the evaluation and performance testing of various XR algorithms, including but not limited to facial expression recognition and reconstruction, facial reenactment, and volumetric video. HEADSET and its all associated raw data and license agreement will be publicly available for research purposes.
}

\keywords{Extended reality, multimodal dataset, virtual reality, volumetric video, light field}

\teaser{
  \centering
  \includegraphics[width=0.9\linewidth]{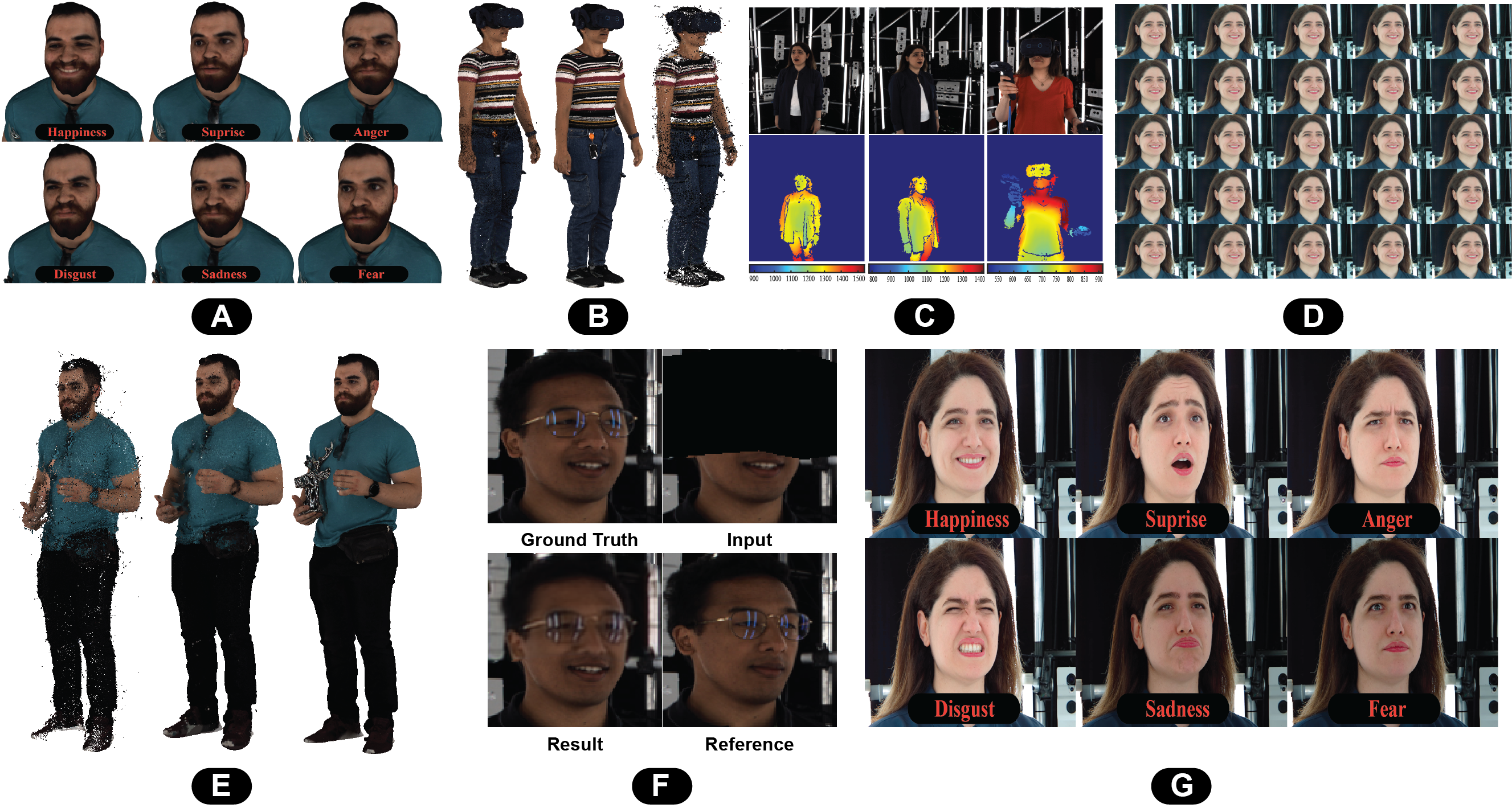}
  \caption{%
  	Details of our proposed dataset (HEADSET) in terms of modalities and use cases. (A): Textured 3D meshes with 6 facial expressions, (B): three types of 3D point cloud with participant wearing an HMD, (C): RGB images and depth maps from 3 views, (D): multi-view representation of LF, (E): point clouds evaluation results, (F): result of HMD removal, and (G): classified LF images with 6 facial expressions. %
  }
  \label{fig:Apps}
}

\renewcommand{\manuscriptnotetxt}{}

\graphicspath{{figs/}{figures/}{pictures/}{images/}{./}} 

\usepackage{tabu}                      
\usepackage{booktabs}                  
\usepackage{lipsum}                    
\usepackage{mwe}                       
\usepackage{graphicx}
\usepackage{url}
\usepackage{paralist,tabularx}
\usepackage{amsfonts}
\usepackage{makecell}
\usepackage{multirow}
\usepackage{transparent}
\usepackage{booktabs}
\usepackage{float}
\usepackage{amsmath}
\usepackage{enumitem} 

\begin{document}


\firstsection{Introduction}
\maketitle
Nowadays, immersive visual technologies including Virtual Reality (VR), Augmented Reality (AR) and Mixed Reality (MR), or short Extended Reality (XR), play a key role in providing virtual experiences for users in various domains, such as XR teleconferencing, XR games, and XR experiences. Photorealistic representation of human interaction is essential for creating a life-like user experience and natural non-verbal communication among users. Specifically, the realistic representation of human facial expressions has a considerable impact on the quality of human interaction and communication ~\cite{kyrlitsias2022social}.

An explicit volumetric representation can be visualized in XR applications by means of colored point clouds or textured meshes. Point clouds have drawn considerable interest due to their relatively simple process of collection and shortage of connectivity information. These features make them appropriate for real-time applications in XR, which require natural communication between users ~\cite{reimat2021cwipc}.

Nonetheless, the lack of human datasets, that are compulsory for volumetric representations, causes many challenges in developing photorealistic XR applications. The development of XR technologies and computer vision techniques is highly dependent on the quality of the datasets that can foster the progress of those fields and their evaluation. In contrast to the high number of research works in designing volumetric data generation algorithms ~\cite{newcombe2015dynamicfusion,alexiadis2016integrated,dou2016fusion4d,orts2016holoportation,jackson20183d,guo2017real,alldieck2018video,guo2019relightables,su2020robustfusion} and three-dimensional (3D) reconstruction ~\cite{saito2020pifuhd,peng2021neural,park2021nerfies}, there have not been enough studies in which we can discover relatively diverse and high-quality volumetric datasets with multiple modalities. 

In this paper, we introduce a new multimodal database that consists of colored 3D point clouds, textured 3D meshes, light field (LF) images, and multi-view RGB-D images acquired from 27 ethnically diverse human participants. We have designed data collection tasks aimed at capturing all participants' posed and spontaneous facial expressions, and body language while speaking. Besides, we have also conducted an experiment for human full-body recording under occlusion caused by an HMD to collect human facial expressions under real-world XR scenarios from 11 participants out of 27. The database captured by our VoCap studio contains 31 ground truth depth cameras and 62 RGB cameras configured in 31 synchronized modules to also allow depth-from-stereo estimation. Additionally, we have used a Lytro Illum camera to collect the human facial expressions of all individuals from a frontal view to address the lack of publicly available LF face resources. Along with the raw and post-processed data, we also provide the labels of the facial expressions for the data collected from the VoCap studio (HEADSET-VoCap) and Lytro Illum LF camera (HEADSET-LF) as part of the main database.

The motivation to create such a dataset is to serve as the basis for research in XR-related use cases, especially in XR teleconferencing, where participants meet and interact in a virtual shared environment. In XR teleconferencing, participants are usually wearing headsets that need to be removed to enable eye contact. The knowledge about the emotion of the participant, who is wearing a headset, might increase the quality of the facial reconstruction. Many studies have focused on HMD removal~\cite{zhao2018identity, wang2019faithful,numan2021generative, chen20223d,lou2019realistic}, which is referred to as the task of virtual removal of HMD, which fill in the occluded color and geometric information of a user’s face. 
The emergence of new MR glasses with emotion recognition capabilities and transparent displays, such as Meta Quest Pro and Apple Vision Pro, may increase the quality of the facial reconstruction results when removing the headset in such a study. 
HEADSET aims at providing ground-truth 3D models of individuals with and without wearing an HMD, and the HMD as an individual object captured with a volumetric capture studio. HEADSET can be utilized by further studies focused on reconstructing faces and gaze directions of the participants under the partial occlusions of the HMD. In this way, the data can be used to evaluate the person's identity after an occlusion removal algorithm is applied.
Moreover, in XR teleconferencing scenarios, volumetric data needs to be compressed and streamed in high quality and low latency to increase the feeling of presence within an immersive environment. As the volumetric capture studio used in this study has the live-streaming capability, it allows XR teleconferencing in real time.  
On top of that, the proposed dataset has potential applications in rendering technologies, animation and simulation, perception, interaction, and user interfaces. We aim to contribute to the development of such approaches beyond their current capabilities to encourage a larger technical advancement in the fast-moving human-centric research domains.

This database can be used as a foundation for testing and validation of various computer vision problems such as 3D face and expression modeling, human activity and movement recognition, multi-view facial emotion recognition, facial reenactment, stereo matching, 3D compression, etc. on real-world high-quality data. In the design process of our data collection tasks and their contents, we have mainly focused on including HMD occlusions and human facial expressions during capturing to address issues related to XR communications. For example, HMDs significantly hinder the virtual experience as the headset covers the person's upper face and eyes. Our dataset thus includes representations of the individuals with and without wearing a VR headset, and the headset as an individual object for studies focused on reconstructing the faces of the participants under HMDs occlusions, as shown in ~\cref{fig:Apps}. Research problems regarding the performance evaluation of facial expression recognition in multi-view RGB images and creating 3D models from a single image are also considered as the purpose of the usage of the proposed dataset. We have therefore incorporated these modalities as part of our dataset, an example of which is displayed in ~\cref{fig:Apps}.
 

The main contributions of this paper can be summarized as follows.

\textit{We introduce a multimodal high-resolution database for immersive media productions in which LF images, RGB images, depth maps, textured meshes, and colored point clouds are crucial. HEADSET and its all associated raw data and license agreement will be publicly available for research purposes\footnote{\url{https://webpages.tuni.fi/headset}}.}

\textit{We collected the data taking into account the diversity of ethnicity and gender from our participants.}

\textit{To the best of our knowledge, we are the first to provide volumetric data as a foundation for applications of emotion and face recognition under partial occlusions. This is done by capturing data of the individuals with and without an HMD to serve as ground truth for real-world XR scenarios.}

\textit{Among many use cases, we selected three applications, in particular multi-view facial expression classification, HMDs removal, and visual quality assessment, for evaluating the dataset. The visual quality experiments are provided on different volumetric representations, i.e. textured meshes and point clouds.}

The remainder of the paper is organized as follows. We first, review the related work in terms of available datasets in ~\cref{sec:related_work}. Then, we present data acquisition steps and the capturing setup in ~\cref{sec:method}. Participant selection criteria and ethical issues are also discussed in this section. ~\cref{sec:experiment} describes the data collection design in our user study. The data post-processing is explained in ~\cref{sec:postprocessing}. We then report and discuss the results of three use cases of HEADSET. Finally, ~\cref{sec:conclusion} summarizes our work.

\section{Related Work}
\label{sec:related_work}

This section reviews related work on available human datasets for each modality, i.e. volumetric data, light field data, and RGB-D data with respect to human participants.

\subsection{Volumetric dataset}

CMUPanoptic~\cite{joo2015panoptic} is the largest public volumetric dataset in terms of the number of capturing modules. In their work, human interactions of 8 participants in distinct social activities were recorded. The multi-view Panoptic Studio~\cite{joo2015panoptic} consists of 31 HD, 480 VGA, and 10 RGB-D (Kinect v2) modules. Although CMUPanoptic stands as one of the largest currently available datasets, it does not provide hardware volumetric synchronization as the time alignment between the Kinect v2 RGBD streams is performed via a hardware modification. 
Zhixuan et al. released HUMBI ~\cite{yu2020humbi}, another publicly available and relatively large multi-view dataset. They captured the human body poses of 772 participants that participated in their study while wearing everyday clothes. The capturing setup included 107 synchronized HD cameras without any depth sensors. Therefore, no information about the ground truth geometry of the scene was acquired. 
Human4D~\cite{chatzitofis2020human4d} is another multimodal, marker-based approach to generate 4D data from volumetric sensors of 4 individuals performing 19 human daily activities. With the purpose of creating a dataset of high movement precision for the development of spatiotemporally aligned poses research, this dataset uses professional motion-capture (MoCap) markers and hardware synchronization for the multi-view data. However, this pursuit for high accuracy has its drawbacks: to produce more authentic movements, only 4 professional actors were selected to produce the 19 scenes, which is detrimental to the diversity aspect of the dataset. Also, the usage of MoCap apparel instead of natural clothing by the participants hinders the potential modeling of gaze, face, and body features. 



Furthermore, one of the latest available dynamic point cloud datasets is CWIPC-SXR ~\cite{reimat2021cwipc}. In this study, 23 human individuals performed activities in social XR settings, thus generating 45 unique scenes. However, the limited number of views used in the capture process -- 7 Azure Kinect DK sensors -- produced low complexity representations, resulting in low-resolution and non-watertight point clouds. In other words, the limited number of views during capture generated models with not so accurate textures and holes in the geometry from the occluded points.


\begin{table*}[ht]
  \centering
  \caption{Comparison of state-of-the-art volumetric datasets with ours in terms of modalities and capturing details.}
  \begin{tabular}{p{2cm} p{1.5cm} p{5.5cm} p{3.5cm} p{3cm}}
    \hline
    \textbf{Dataset} & \textbf{Participants} & \textbf{Description} & \textbf{Data type} & \textbf{Capturing modules} \\ \hline
    CMUPanoptic ~\cite{joo2015panoptic} & 8 & Participants performing social activities & Multi-view RGB-D, 3D point clouds & 31 HD, 480 VGA, and 10 RGB-D \\ \hline
    HUMBI ~\cite{yu2020humbi} & 772 & Human body expressions & Multi-view RGB, 3D meshes & 107 HD \\ \hline
    Human4D ~\cite{chatzitofis2020human4d} & 4 & Professional actors performing full-body movements and expressions & Multi-view RGB-D, point clouds, 3D meshes & 24 motion capture and 4 depth sensors\\ \hline
    CWIPC-SXR ~\cite{reimat2021cwipc} & 23 & Human interaction in social XR settings & Multi-view RGB-D, 3D point clouds & 7 Azure Kinect DK sensors \\ \hline
    8iVSLF ~\cite{krivokuca20188i} & 6 & Full-body of a human participant & 3D point clouds & 39 RGB cameras in 12/13 rigs \\ \hline
    Volograms\&V-SENSE ~\cite{pages2021volograms} & 3 & Monologue and dancing & Meshes with texture images & 12/60 camera studio \\ \hline
    \textbf{HEADSET} & 27 & Posed facial expressions and subtle body movements w/o VR headset & Multi-view RGB-D, 3D point clouds, 3D textured meshes, light field & 62 RGB, 31 depth, and 1 Lytro Illum cameras \\ \hline
  \end{tabular}
  
  \label{tab:literature}
\end{table*}

The 8iVSLF dataset ~\cite{krivokuca20188i} is another dynamic voxelized point cloud dataset only containing 6 high-resolution single-frame models for 6 human individuals. The capturing setup included 39 synchronized RGB cameras configured in either 12 or 13 rigs. 

Finally, Volograms \& V-SENSE ~\cite{pages2021volograms} have also introduced a small volumetric video dataset. The published dataset consists of three textured mesh sequences with differing characteristics and relatively short sessions. The dataset was captured in VoCap studios, which include 12 or 60 studio cameras.

In addition to the available volumetric datasets, HEADSET enables further research perspectives by collecting multimodal human data with the help of numerous camera modules. In our proposed dataset, along with the LF data, we present new volumetric sequences (HEADSET-VoCap) captured by a VoCap studio including 62 RGB and 31 depth cameras arranged in 31 synchronized camera modules. It contains post-processed textured meshes, point clouds with different resolutions, and multi-view RGB-D frames from 27 individuals displaying posed facial expressions and subtle body movements while speaking. Sequences under HMD occlusions are also part of the main database to introduce additional modalities compared to the available volumetric datasets.
To better compare the proposed database with existing volumetric human databases as described before, we give an overview in ~\cref{tab:literature}.   

\subsection{Light field dataset}
To the best of our knowledge, only four LF human face datasets have been made publicly available. 
The Light Field Face and Iris Database (LiFFID) ~\cite{raghavendra2015exploring} is the first human face dataset that contains images captured with an LF camera for the purpose of facial recognition. It comprises a group of 2D greyscale images created from the LF content captured by a Lytro lenslet camera. Nevertheless, LiFFID does not contain raw LF images, which is a considerable challenge for many research fields.
The IST-EURECOM Light Field Face Database (LFFD) ~\cite{sepas2017eurecom} is the second LF face dataset which includes both raw and rendered data from 100 persons, with 20 LF samples per participant acquired by a Lytro Illum lenslet camera in a controlled capturing setup with several facial variants.
In ~\cite{sepas2021capsfield}, the authors introduced the Light Field Faces in the Wild (LFFW) and Light Field Face Constrained (LFFC) face datasets. LFFW includes 1908 LF images from 53 individuals captured in the wild in both indoor and outdoor environments without any predefined protocol. LFFC complements the LFFW dataset by including 1060 LFs from the same 53 participants acquired in constrained conditions.
Despite the recent advances in LF face analysis ~\cite{galdi2019light} and facial expression recognition ~\cite{sepas2021capsfield}, highly accurate recognition results are still not achievable for some specific conditions due to the lack of data. 

In our work, we present the HEADSET-LF dataset in addition to the volumetric data, which contains two subsets. The first one is collected from 27 participants showing 6 basic human emotions, totaling 162 LF images with corresponding labels for the facial expressions. The second one includes 10 LF frontal images of 10 individuals wearing a VR headset. Since this dataset will be publicly available, it may be used as the basis for the future validation and assessment of LF-based facial expression classification and recognition as well as facial reconstruction.

\subsection{RGB-D dataset}

Despite the wide availability of large RGB face image datasets ~\cite{mollahosseini2017affectnet,li2017reliable,barsoum2016training}, similarly sized datasets containing RGB-D face images are not available yet. RGB-D face datasets contain a limited number of samples ~\cite{chhokra2018unconstrained, min2014kinectfacedb, zhang2016lock3dface}, or they have been captured without considering HMD occlusions ~\cite{zheng2022complementary} and any additional modalities.
Hence, researchers mostly tend to use a synthetic dataset with a high degree of variety in order to solve their research problems related to human faces.
While the usage of a synthetic dataset reduces the potential of generalization to real-world data, this approach has been widely used in the literature. For example, for HMD removal/ facial reconstruction, the authors of ~\cite{numan2021generative} built a data synthesis pipeline to create a synthetic dataset of RGB-D images with a random pose, ambient illumination, and expression of faces based on the Basel Face Model (BFM) 2017 ~\cite{gerig2018morphable}. 
To address the aforementioned challenges, our dataset contains additional sequences under HMD occlusions in order to be used as the testing set for the future validation and assessment of such research work. Along with LF and RGB-D data, HEADSET also includes 3D point clouds and 3D textured meshes with an average number of frames of 272 for every 27 individuals. The sequences under HMD occlusions also contain 58 frames per participant for 11 participants out of 27.

\section{Data Acquisition}
\label{sec:method}

In this section, we provide more details of the data acquisition process with regard to the capturing setups, VoCap studio and Lytro Illum camera, and the performed steps for participant selection.

\subsection{Capturing Setup} 
\label{subsec:setup}

\subsubsection{Volumetric capture studio}

\begin{figure}
    \centering
    \includegraphics[width=1\linewidth]{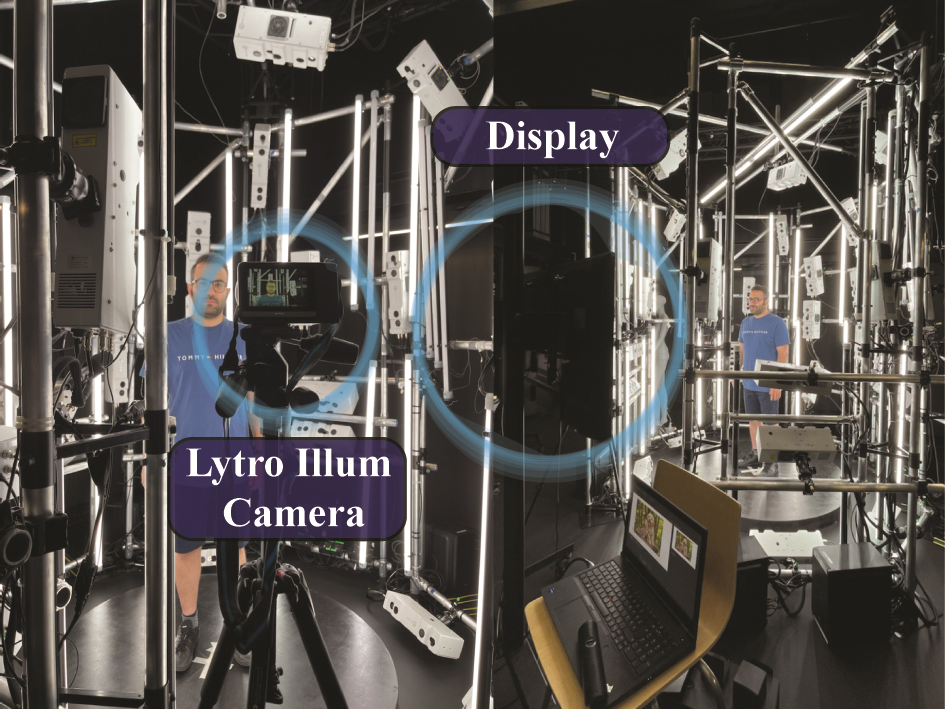}
    \caption{The complete capturing setup for the data collection. In addition to the VoCap studio, we also used one Lytro Illum camera, one display, and a microphone.}
    \label{fig:lytro}
\end{figure}

We have utilized a 3D VoCap Studio (Mantis Vision Volumetric Capture System\footnote{\url{https://mantis-vision.com/3d-studio-3iosk/}}, version: studio ring) for capturing the volumetric dataset. A custom room setting with a cylindric recording area was employed (radius: 1.6 m, height: 2.5 m). Similarly, the capture rigs were placed in a cylinder with radius of 2.5 m and height of 3 m. The VoCap studio contains an aluminum frame with a black background and adjustable lighting. Hence, the capturing scene was illuminated by $34 \times$ Quasar Science Q50XG lights. They were spread evenly around the studio in order to provide enough light for our capturing scenes. The floor of the recording space is also black. The black backgrounds and floor reduce reflections during the capturing process. The complete setup is shown in ~\cref{fig:lytro}. The studio has three types of uEye cameras and 31 camera modules. Each camera module has a laser and monochromatic UI-3140xCP-M for structured light based depth estimation. In addition, it has two UI-3080xCP-C or UI-3280xCP-C cameras for color information. Therefore, a total of 62 RGB cameras ($2054\times2456$ pixel) and 31 depth cameras ($1024\times1280$ pixel) were used. 

The depth camera supports several modes by which the frame rate, resolution, exposure time, operating range of the module, and region of interest can be modified. The modes of the color camera, including frame rate, resolution, field of view, aspect ratio, and format can be also determined. In our work, all the raw data had been captured with 25 frames per second (fps) during the data collection. 


The calibration and synchronization of the VoCap studio was carried out once before starting the capture, and the calibration parameters are enclosed within each dataset based on its modality. As the VoCap studio can be seen as a single fully calibrated capturing device for capturing high-quality 3D models, the recorded volumetric data can be used as ground truth. However, we extracted 3 RGB-D module outputs separately based on their field of view and our experiments' analysis. Thus, the exact internal and external camera parameters have also been provided within the RGB-D dataset. When the participant stands in the center, the distance to the cameras varies roughly from 80 cm to 120 cm. While standing in the center and looking ahead, 11-13 camera modules have a good view of the person's face within about 130 degrees angle in the individual's field of view.

\subsubsection{Lytro Illum light field camera} 

Complementary to the VoCap studio, we added one microphone to record the audio signal, one frontal Lytro Illum LF camera to capture the frontal view, and one large display for showing the necessary content of the experiment to the participant. The Lytro Illum is a 40 megaray LF camera with constant F/2.0, 8X optical zoom, and a 4-inch tilt screen. The camera provides data in Light Field Raw (LFR) and depth map (PNG and TIFF, $2022\times1404$ pixel) file formats, which are contained in our dataset. The LFR file stores the image information in an uncompressed lenslet image before further processing. The relative depth of field coordinates ($\lambda_{max}$ and $\lambda_{min}$) along with the calibration information have been also included in the dataset, which has been extracted using the Lytro Desktop Application. The camera height and zoom ratio were properly adjusted to capture the individual's face based on its height. During the capturing process, the camera was located in front of the participant while the person was standing at the VoCap studio’s central point (153 cm distance from the Lytro camera’s location). Calibration was performed before every recording session, using a 3 m, 16 mm steel tape measure. The complete setup and the location of the display, and LF camera are depicted in ~\cref{fig:lytro}. 

\subsection{Participant selection}

\subsubsection{Ethical issues}
\label{subsubsec:ethic}

Before taking part in the data collection tasks, each individual had to read and sign an information sheet and consent form, which allowed the use of data for research purposes and data publication.  

The participants were also supposed to read the safety training material. All of the possible risks of taking part in our study, such as physical discomfort, which may potentially happen due to wearing a VR headset or posing facial expressions, have been mentioned in the participant information sheet. In order to collect high-fidelity records, the volunteer was asked to avoid large head and body movements during the capturing. The ethical issues in our work had been carefully considered and were fully approved by the Academic Ethics Committee of the Tampere Region, Tampere University. The complementary explanations of the ethics procedure in our work can be found in the supplementary material. 

\subsubsection{Participants}
\label{subsubsec:participants}

We looked for people who were interested to take part as volunteers in our study. There were no specific criteria for participant inclusion, with the exception of each participant having to be above the age of 18 years. 
Although we did not inquire about the participants' ethnicity due to ethical concerns, we attempted to keep the dataset ethnically diverse to the best of our efforts. 
There are 19 male participants with an average age of 26.37 years and an average height of 178.05 cm. The corresponding figures for the 8 female participants are 27.5 years and 165 cm. 10 out of 27 participants were wearing glasses at the time of the capture, there being 9 male and 1 female.  

\section{Data collection design}
\label{sec:experiment}

Our data collection tasks aimed to capture posed expressions (task A), spontaneous facial expressions, and body poses while speaking (task B). Moreover, we have also collected human face and full-body recording under occlusion caused with a VR headset (task C), i.e. we have designed three types of assignments for data collection. 

Before starting the data collection, two training examples were given to the volunteers to explain what they were supposed to do during each session: one before A and B, and the other before C. Thus, tasks A and B were carried out uninterrupted. In the first training session, we explained the tasks of tasks A and B to the participants and showed them sample images (dissimilar to the images shown in the effective tasks but in the same category). In the second training session, which was performed after finishing assignments A and B, we asked the participant to wear a VR headset. More details of the data collection task are explained in the following subsections. 

\subsection{Data collection Task A}

This data collection task consisted of showing the volunteer 6 basic human emotions on a big display screen in front of the person and the VoCap studio. The basic emotions included \textit{happiness, surprise, anger, disgust, sadness,} and \textit{fear}. The target emotions were defined and displayed as described in ~\cite{verpaalen2019validating}. Presenting facial images on a display was our main way to elicit such expressions. The participants were asked to look at each picture that appeared on the display and then to try to mimic the expression that had the same semantic meaning as the displayed one.



\subsection{Data collection Task B}

In this assignment, we displayed three pictures to the volunteer with background sounds related to the content of the pictures. These three images contained animals, a baby, and a nature scene, respectively. The participants were then asked to look at each picture and describe it and their feelings about each of them in their own words. The spoken language was either English or the participant’s mother tongue based on their own preference. We encouraged them to express their thoughts in their mother tongue so that they could generate facial expressions that were as close to natural as possible when speaking. In total, 15 out of 27 participants spoke in English, and the rest preferred to speak in their mother tongue.   


\subsection{Data collection Task C}

In the final task, the participants were asked to wear an HTC Vive Pro Eye headset ~\cite{ViveProEye} while standing in the center of the capturing studio, where they looked at different points on the headset’s display. The participants were supposed to move their body around without changing their location. During this process, participants mostly showed Neutral emotions while looking at different cameras.    
In addition to recording sequences from the volunteers, we also reconstructed a 3D model of the VR headset, which we used for task C, for 2 frames. This recording aims at providing a ground-truth 3D model of the occlusion object for further studies focused on reconstructing faces and gaze directions of the participants under the partial occlusions of the headset. In cases where identity preservation is critical, such as HMD removal and facial expression reconstruction, we have provided representations of the individual wearing a VR headset and without it, and the headset as an individual object. In this way, all three types of data can be used to evaluate the identity of the person after an occlusion removal algorithm is applied.

\section{Data post-processing}
\label{sec:postprocessing}

We recorded over 5 hours of raw data with 25 frames per second (fps). However, it was neither feasible nor necessary to post-process all of them mainly because of the high computational cost and memory issues. Therefore, we post-processed each of the sequences based on data type usage with different segments. The post-processing frame rate was 3 fps for data collection tasks A, B, and C. 
It is worth mentioning that in addition to the post-proceed data, the raw captured data @25 fps and camera calibration parameters are also made available. Each data type has been organized according to the participant's identification number and frame number. Detailed instructions on data structure and synchronization information are also given within each dataset.  
In this section, we go through each data type and explain the applied post-processing steps.

\subsection{Textured 3D meshes}
In order to reduce the amount of data processing as well as to avoid collecting too many similar frames, we decided to reduce the sampling rate of the post-processed data. In order to generate the 3D meshes, the Poisson surface reconstruction \cite{kazhdan2020poisson}technique was applied from the raw point cloud data. In addition to capturing sequences from the participants, we also built a 3D model of the VR headset, which we used for data collection task C, for 2 frames. 
\begin{figure}[t]
\centering
\includegraphics[width=0.8\linewidth]{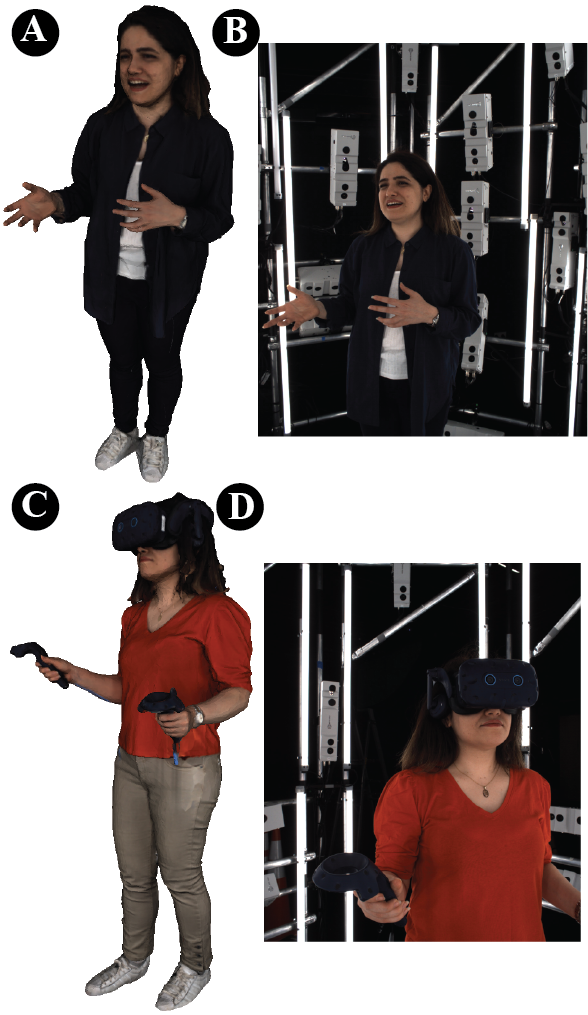}
\caption{Example of reconstructed textured meshes. (A): full-body 3D model of a participant, (B): RGB image captured by camera number 30, (C): full-body 3D model with HMD occlusion, and (D): RGB image captured by camera number 16.}
\label{fig:3D}
\end{figure}
~\cref{fig:3D} illustrates examples of reconstructed textured meshes post-processed after recording, and RGB images captured during the recording of data collection tasks B and C. 

\subsection{Colored 3D point clouds} 
\label{sub:colored-pcs}

\begin{figure}[t]
\centering
\includegraphics[width=0.9\linewidth]{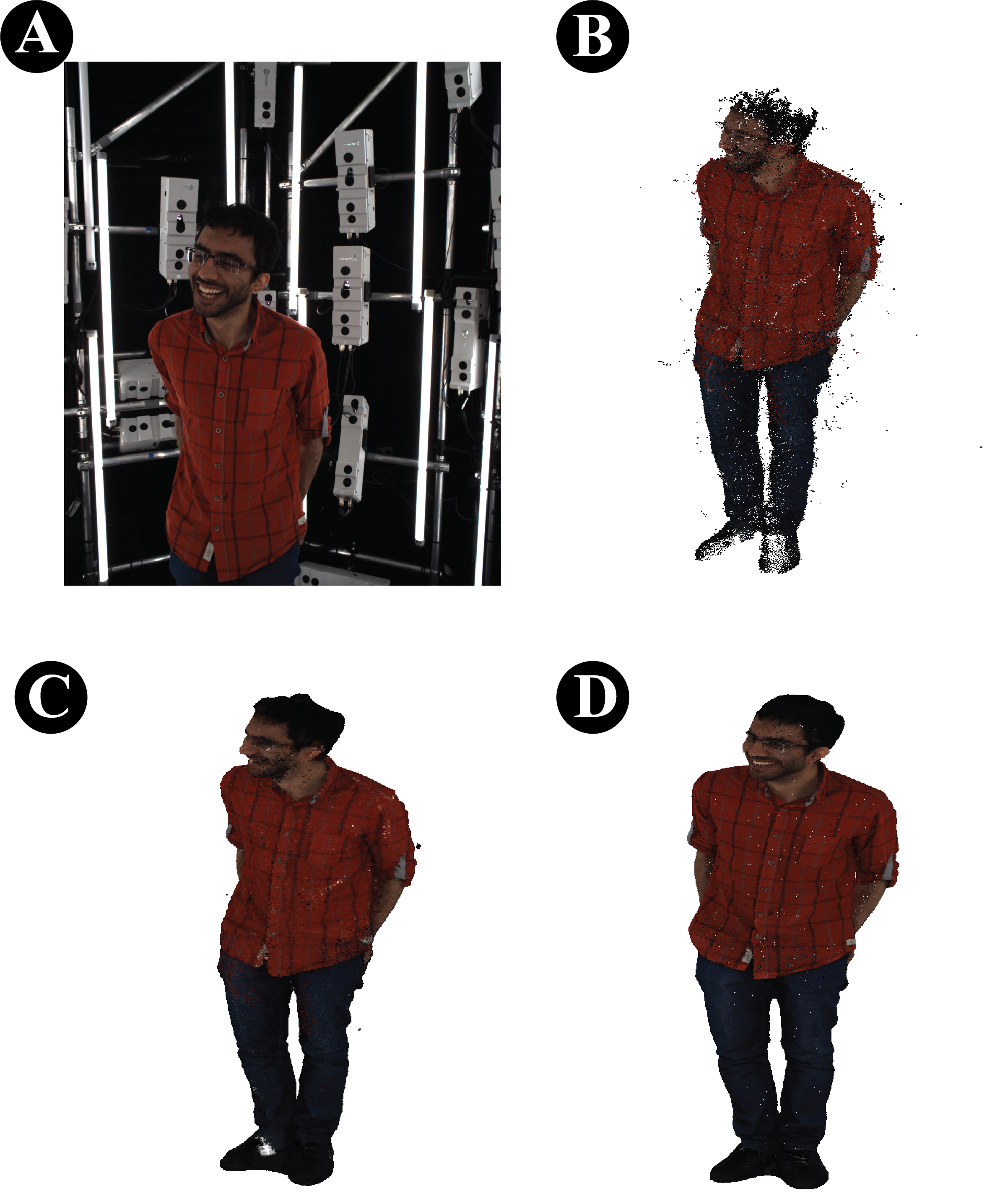}
\caption{Example of colored point clouds of a participant wearing glasses in task A. (A): RGB image captured by camera number 30, point cloud representation from (B): raw data, (C): post-processed, and (D): sampled from textured meshes.}
\label{fig:PC}
\end{figure}

The raw point cloud data is obtained by generating the geometry from the ground truth depth maps captured by the 31 depth cameras of the VoCap Studio. The 2 stereo images from each capture module further improve the geometry by applying depth-from-stereo, while also coloring the scene's points. ~\cref{fig:PC} illustrates examples of reconstructed point clouds acquired after recording and corresponding RGB images captured from an individual wearing glasses during the capturing process.

Due to the sparse and noisy nature of the raw point clouds (~\cref{fig:PC}-B) -- which contain around $\sim300,000$ points --, we also provide post-processed versions of them. This is done by removing outlier points and applying a Poisson surface reconstruction \cite{kazhdan2020poisson}, as done in \cite{girardeau2016cloudcompare}, to increase their resolution. Afterwards, we sample the points from the mesh \cite{metro_sampling} with a surface density -- the number of points per square unit -- of 0.05, resulting in point clouds of around $\sim900,000$ points. One example is depicted in ~\cref{fig:PC}-C. Finally, even though the post-processed point clouds increase the models' resolution, certain materials like extremely non-Lambertian objects, \textit{e.g.} mirror-like surfaces, are not well represented only from the RGB-D images due to the lack of geometry of the raw depth data. Thus, we also provide an additional type of post-processed point cloud, which is sampled from textured meshes. An example of this type of a post-processed point cloud is illustrated in ~\cref{fig:PC}-D.
 

\subsection{RGB-D}
\label{subsec:RGBD}
We have also collected RGB-D images for tasks A and B from two of the capture modules based on the cameras' field of view and the capture setup's layout. Their corresponding indices are 1 and 30. These indexes were chosen since they provided a good field of view for capturing the volunteer’s facial expressions. The distance between the depth cameras and between the RGB cameras of modules number 30 and 1 is 989.565 mm and 991.126 mm, respectively.
At the beginning of each sequence, the participants were trying to understand the first task. To that end, we decided to check all the sequences manually and remove the redundant frames from the start and end of capturing to avoid the collection of many similar and incomplete frames. The script for exporting the depth maps for each frame is also published together with the RGB-D dataset. The script that reads, processes, and visualizes the camera transformations is included in the dataset along with the exact positions of each camera. Therefore, each subset of the RGB-D data processed for all data collection tasks contains the extrinsic and intrinsic calibration matrices for both RGB cameras and the extrinsic matrix for each module's depth sensor at the time of capture. 

For task C in which the person was wearing a VR headset, RGB-D images have been collected from one frontal module (number 16) for 20 seconds at 3 fps. Here, module number 16 provided the frontal view because the participant was looking in the opposite direction compared to data collection tasks A and B. 

\begin{figure}
\centering
\includegraphics[width=1\linewidth]{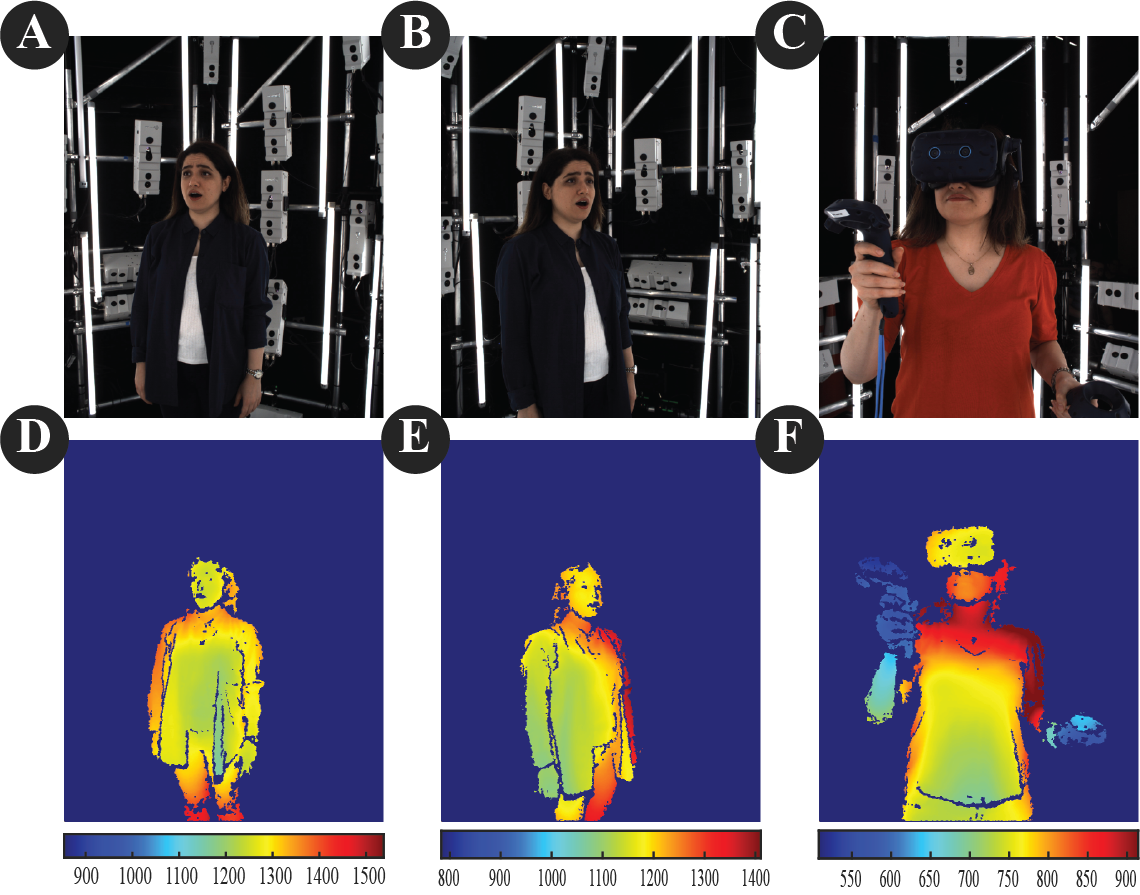}
\caption{Sample of RGB images and depth maps from three views. (A,D): RGB-D image of module number 30, (B,E): RGB-D image of module number 1, (C,F): RGB-D image of module number 16.}
\vspace{-2ex}
\label{fig:depth}
\end{figure}

In ~\cref{fig:depth}, a sample of RGB images and depth maps from camera number 1 and 30 is shown, where the individual was performing the "Surprise" expression. An example of RGB-D representation from module 16 in task C, in which the volunteer is wearing an HMD, is also depicted in ~\cref{fig:depth} (C, F).

\begin{table*}
  \scriptsize
  \centering
  \caption{HEADSET modality details with regard to its content. All the raw data captured by the VoCap studio @25fps is additionally available in .mvx file format with a total size of 2598 GB.}
  \resizebox{1\textwidth}{!}{%
  \begin{tabular}{p{1.5cm} p{1.5cm}  p{3cm} p{2.5cm} p{1cm} p{1.2cm} p{1cm} p{1cm}}
    \toprule
    \textbf{Data type} & \textbf{Participants (out of 27)} & \textbf{Content} & \textbf{Occlusion} & \textbf{Capturing modules} & \textbf{Avg. \# of frames} & \textbf{Avg. size}\ & \textbf{Format}\\ 
    \hline
    Colored point clouds & 27 & 6 posed expressions and subtle body movements & Natural (caused by glasses or hair) & 31 & 272 & 4.33 GB & .ply\\ \hline
    Colored point clouds & 11 & Subtle body movements & HMD & 31 & 58 & 885 MB & .ply\\ \hline
    Meshes with textures & 27 & 6 posed expressions and subtle body movements & Natural & 31 & 272 & 2.07 GB & .obj\\ \hline
    Meshes with textures & 11 & Subtle body movements & HMD & 31 & 58 & 339 MB & .obj\\ \hline
    RGB-D & 27 & Posed and spontaneous facial expressions & Natural & 2 & 455 & 6.7 GB & .png\\ \hline
    RGB-D & 11 & Subtle facial movements & HMD & 1 & 20 & 595 MB & .png\\
    \hline
    LF & 27 & 6 posed expressions & Natural & 1 & 6 & 377 MB & .lfr\\
    \hline
    LF & 10 & Face & HMD & 1 & 1 & 65 MB & .lfr\\
    \bottomrule
  \end{tabular}
}

  \label{tab:data}
\end{table*}

\subsection{Light field}

The LF image dataset consists of two subsets according to their content. The first one was collected from 27 volunteers showing 6 basic human emotions, totaling 162 LF images. The data is labeled into 6 classes based on the participant’s performed emotion. The second one includes 10 LF frontal images of 10 people wearing an HMD.
Along with LF raw data (.lfr files), their related depth maps are also available in TIFF and PNG file formats.

~\cref{tab:data} summarizes the details of the HEADSET multimodal dataset captured by the VoCap studio and Lytro Illum camera.


\subsection{Post-processed labeled data}
\label{subsec:labeledData}

We have also created two RGB labeled subsets of our main database which include human facial expressions. The first one is the multi-view representation of LF data captured by the Lytro Illum camera, and the second one contains RGB images from two non-frontal views captured by the VoCap studio. The label of each image has been defined based on the ground truth emotion described in task A. The images that we used further for the evaluation of the facial expressions classification (FEC) in our dataset include VoCap RGB data (HEADSET-VoCap), and multi-view representation of light field data (HEADSET-LF), both in PNG file format. HEADSET-LF is created from sub-aperture images of the LF raw data as multi-view RGB images. For this work, each LF raw data is converted into a $5\times5$ RGB view matrix.
It is noteworthy that in some cases the participant showed "Neutral" emotion instead of the required emotion. Thus, we removed the samples that are apparently not matched to the ground truth label by human observation. However, we made both the original and the modified datasets with labels available.  
The number of RGB images in the modified datasets, which we used for the evaluation, are as follows: HEADSET-VoCap: \textit{\{Anger: 363, Disgust: 266, Fear: 209, Happiness: 264, Sadness: 284, Surprise: 420}\} and HEADSET-LF: \textit{\{Anger: 650, Disgust: 550, Fear: 450, Happiness: 675, Sadness: 375, Surprise: 575}\}, which are acquired from the multi-view RGB representations of the raw LF images.

\begin{figure}[t]
\vspace{0.8em}
\centering
\includegraphics[width=1\linewidth]{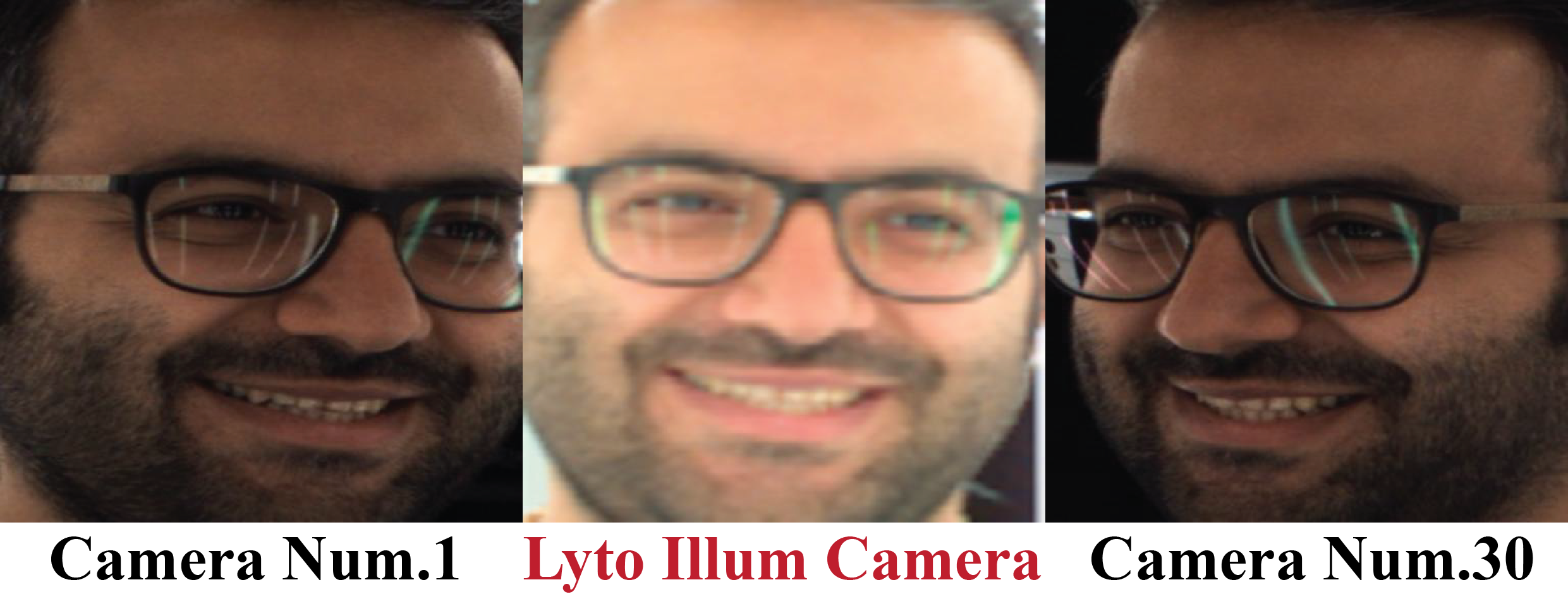}
\caption{Three synchronized views (two non-frontal views in HEADSET-VoCap, and one frontal view in HEADSET-LF) of detected faces showing a "Happiness" expression.}
\label{fig:3view}
\end{figure}

For our evaluations and experiments, which we describe in ~\cref{sec:result}, we first applied a deep cascaded multi-task framework method for face detection (MTCNN) proposed in ~\cite{zhang2016joint} on both labeled datasets in order to make the evaluation faster. We then checked the facial images and landmarks of all views of the dataset and proved that the facial region is detected in all the frames.

~\cref{fig:3view} depicts two non-frontal views (left: camera number 1, and right: camera number 30 of the VoCap studio) of HEADSET-VoCap, and one frontal view (middle: captured by the Lytro Illum camera, central sub-aperture) of HEADSET-LF as examples for the "Happiness" class after applying the face detection algorithm.

\section{Experimental results and dataset evaluation}
\label{sec:result}

We conducted multiple experiments using different types of our data in order to report HEADSET's performance compared to similar ones in a multitude of applications. We first present the volumetric assessment of sequences under the HMD occlusion compared to a similar currently available dataset. Then, we evaluate the dynamic scenes in the context of compression with two state-of-the-art 3D codecs for voxelized point clouds. Afterward, we focus on two popular computer vision problems which can be a use case of the proposed dataset. The first application involves facial expressions classification in HEADSET-VoCap and HEADSET-LF collected from Experiment A, as described in ~\cref{subsec:labeledData} in order to prove the expression variations in the collected data. We also present the results of a deep video inpainting model on our dataset for solving the HMDs removal problem as the second application. In this section, we explain each of the experiments in detail. 

\subsection{No-reference volumetric assessment of headset-wearing participants}

Although the volumetric datasets from \cref{tab:literature} provide different scenes for immersive applications, including typical social XR situations as well as body poses and movements, few of them include a photorealistic representation of the participants' occluded expressions. More precisely, only the CWIPC-SXR dataset \cite{reimat2021cwipc} contains such data from the aforementioned datasets, where two scenes depict three individuals performing actions while wearing an HMD. Our HEADSET data contains volumetric information (both meshes and point clouds) from high-resolution captures for 11 different persons wearing an HMD, while also providing scenes of the same individuals without it. 


In order to assess the quality of our generated volumetric data, we estimate the subjective quality of our 3D data using a no-reference (NR) metric to evaluate the post-processed point clouds derived from the textured meshes as described in \cref{sub:colored-pcs}. The usage here of an NR point cloud quality assessment (PCQA) is paramount due to the lack of a reference point cloud to compare to. That is, potential distortions over the geometry and texture data are due to the nature of the capturing and processing pipeline of the dataset to generate the scenes. Therefore, it is important for the selected NR-PCQA metric to have a good generalization capability over the different kinds of possible distortions regarding the geometry and the attributes. 

With that in mind, we assess our generated data for participants wearing an HMD via the ResSCNN NR metric \cite{liu2023-nrpcqa}, which leverages a sparse convolutional neural network to estimate the subjective quality of point clouds without the usage of reference models. To properly evaluate our and the CWIPC-SXR scenes, we use a ResSCNN model trained on a large-scale point cloud quality assessment dataset, which contains 104 reference point clouds with more than 22,000 example cases with 31 different types of distortion over the geometry and the attributes data. These distorted samples are annotated with a pseudo mean opinion score (MOS) to subjectively evaluate the 3D data (see \cite{liu2023-nrpcqa} for a more detailed explanation). In short, this pseudo MOS is a scale of five quality levels in the range $[1,5]$, where 1 means that the distortions significantly hinder the perception of the scene and 5 means that almost no distortion is perceived.

Our experiments are performed over 20 frames for each of the scenes. The 11 participants of HEADSET-VoCap are evaluated against the two scenes from the CWIPC-SXR \cite{reimat2021cwipc} dataset that contains participants wearing an HMD: scene 14 (``Rock-paper-scissors in VR"), containing two persons, and scene 19 (``Boxer in VR"), with one. As recommended in \cite{liu2023-nrpcqa}, all the sequences' coordinates were scaled in the range of $[0-2000]$ for the evaluation.

\begin{table}[]
\centering
\caption{Average Pseudo Mean Opinion Scores of participants wearing a head-mounted display. Higher is better. Results were averaged from 20 frames of 3 individuals from \cite{reimat2021cwipc} and 11 individuals from our dataset.}
\label{tab:pseudo-mos}
\begin{tabular}{ll}
\hline
\multicolumn{1}{c}{\textbf{Point cloud type}} & \textbf{MOS} $\uparrow$ \\ \hline
CWIPC-SXR \cite{reimat2021cwipc}                            & 2.885        \\
Raw                                 & 4.127        \\ 
Post-processed                                 & 4.443       \\ 
Sampled from texture meshes                                 & \textbf{4.853}        \\ \hline
\end{tabular}
\end{table}

Results from \cref{tab:pseudo-mos} suggest a superior subjective quality of our scenes than the ones in \cite{reimat2021cwipc} according to their pseudo-MOS. In particular, these results show that even our raw types of point clouds still present a decent subjective quality, even in the presence of outliers. Our post-processed scenes, which include the outlier removal in the post-processing step -- a procedure that is also done for the data in \cite{reimat2021cwipc} --, show an even greater improvement over our raw types, which is expected, with the ones derived from the textured meshes performing the best out of them. Moreover, we not only provide a larger number of participants, but also their ground truth non-occluded physiognomies, making our data suitable for applications targeting the study of facial occlusion directly over the 3D data.

\begin{figure}
\centering
\includegraphics[width=0.8\linewidth]{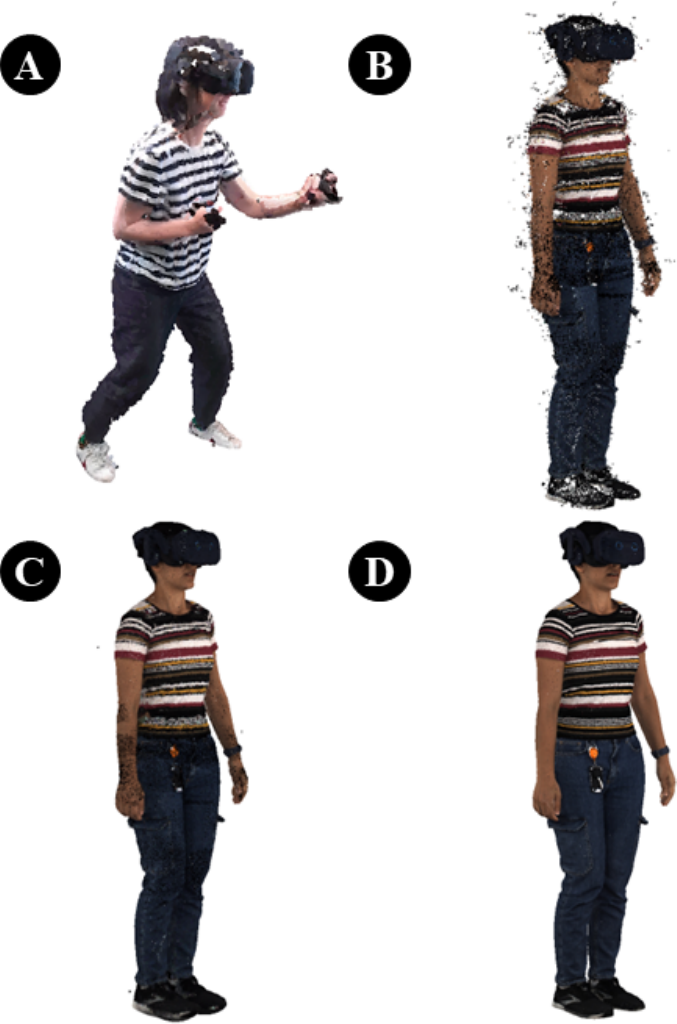}
\caption{Visualization of a frame from a) ``Boxer in VR", from \cite{reimat2021cwipc} and versions for our three types of point clouds: b) raw, c) post-processed, and d) derived from textured meshes, for Participant 19 of our dataset.}
\label{fig:nr-pcqa-visualization}
\end{figure}

\begin{figure}[htb]
\centering
\begin{subfigure}[h]{\columnwidth}
    \includegraphics[width=\linewidth]{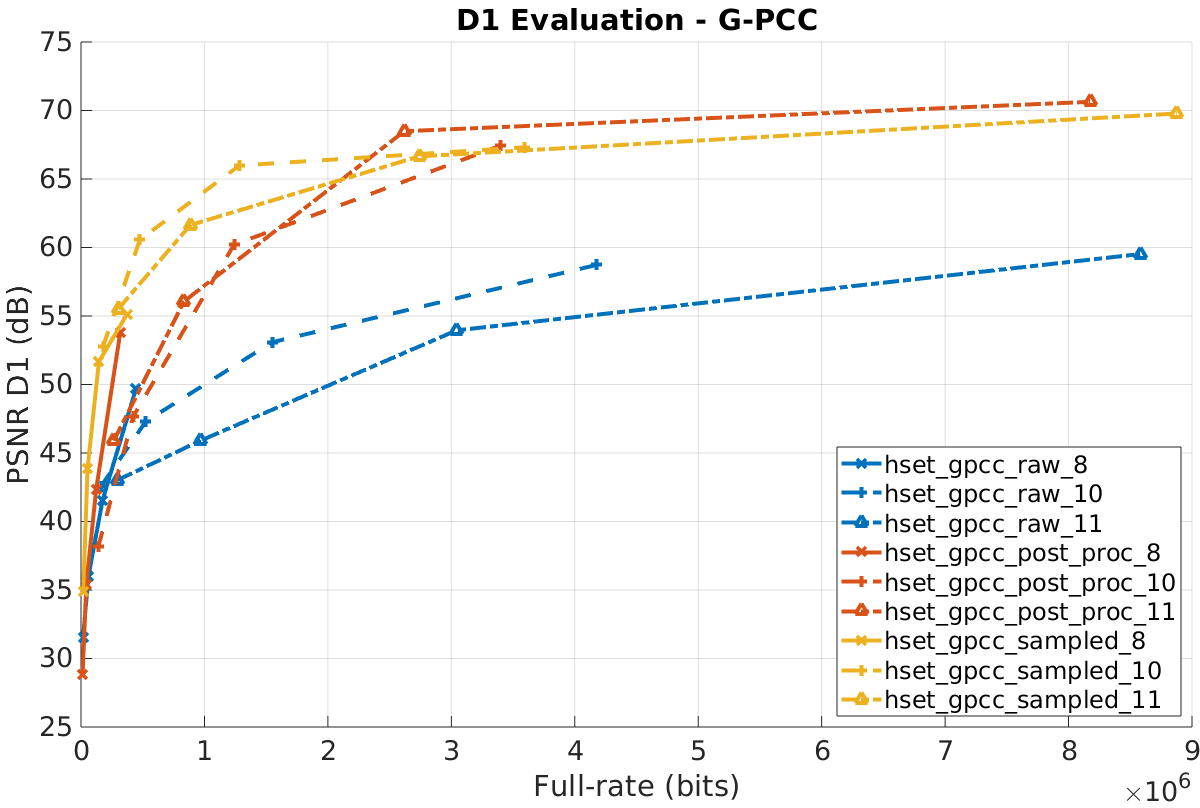}
\end{subfigure}
\begin{subfigure}[h]{\columnwidth}
\includegraphics[width=\linewidth]{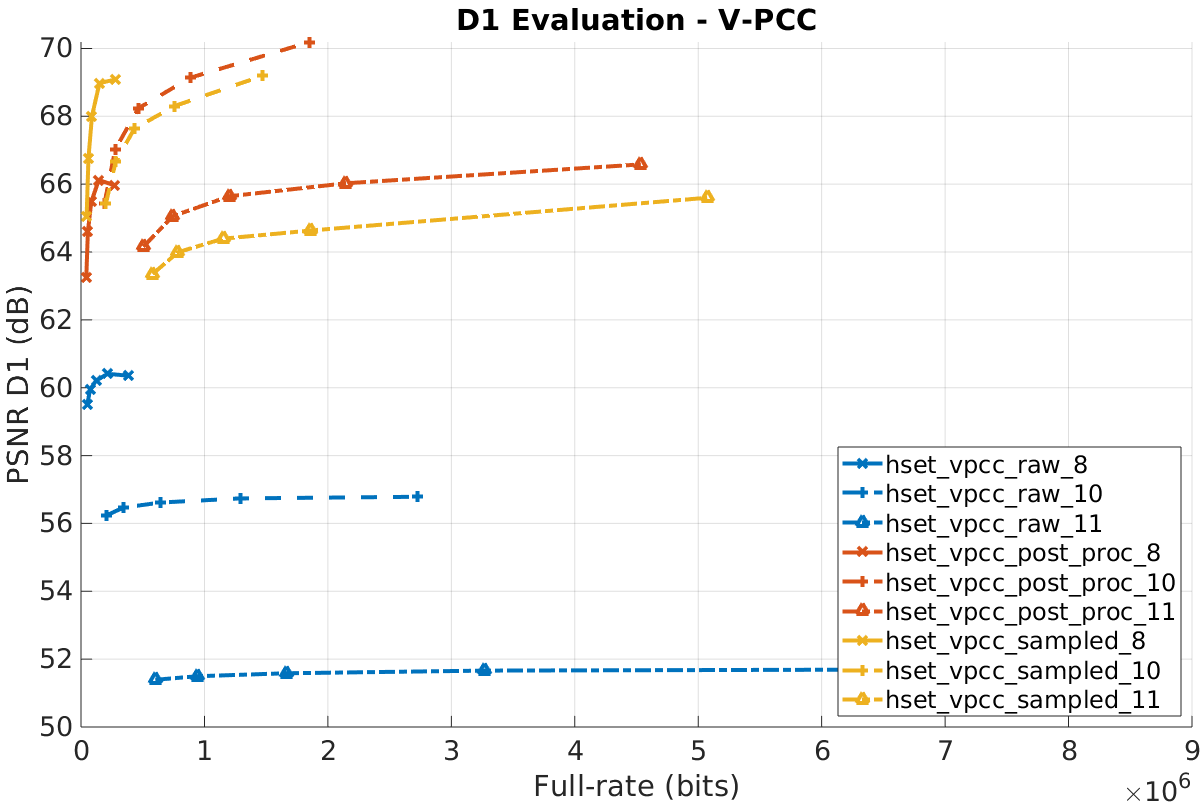}
\end{subfigure}
\caption{RD Results for the average D1 metric from 4 frames for codecs G-PCC (top) and V-PCC (bottom).}
\label{fig:d1-rd-results}
\end{figure}

\cref{fig:nr-pcqa-visualization} also shows an illustration of four 3D models of the compared datasets, with one frame from the sequence ``Boxer in VR" from \cite{reimat2021cwipc} and the other three being our three types of point clouds. Notice how the sequence from \cite{reimat2021cwipc} in \cref{fig:nr-pcqa-visualization}-a) presents significant distortions in the geometry from ``holes", \textit{i.e.} missing points, and also in its colors. On the other hand, the frame taken from the raw point cloud for participant 19 of our dataset in \cref{fig:nr-pcqa-visualization}-b) appears to have less evident distortions for the colors, while presenting a more significant number of outlier points. This number of outliers is greatly reduced for our post-processed sequence in \cref{fig:nr-pcqa-visualization}-c), although it presents some minor holes and lacks some of the finer texture and geometry details. Our 3D model derived from the textured meshes, which can be seen in \cref{fig:nr-pcqa-visualization}-d), not only provides a watertight geometry but is able to depict some of the scene's finer details, such as the watch and the harness on the person's belt. 

\subsection{Point clouds evaluation in terms of compression}

In order to popularize applications that provide virtual experiences with low latency, like XR telepresence, XR games, and free-viewpoint videos, it is paramount for the volumetric data to be efficiently conveyed in real-time. As such, we benchmark our dynamic scenes in the context of compression with two state-of-the-art 3D codecs for voxelized point clouds from the Moving Picture Experts Group (MPEG): the geometry-based point cloud compression (G-PCC) and the video-based point cloud compression (V-PCC) standards \cite{emerging_mpeg}.
Since our dataset consists of sequences with multiple frames of volumetric data for each scene, we provide a testbench for the development of both static -- or intra-frame -- and dynamic -- or inter-frame -- methods for volumetric video compression. Hence, we selected V-PCC due to its suitability for temporal video compression, and G-PCC for its profile of static data compression. More details on both solutions can be found in \cite{emerging_mpeg,overview_gpcc_vpcc}.

As both codecs require the point cloud data to be voxelized, that is, the points are quantized into volumetric elements, we constrain our data into three different voxel grid resolutions, $b = \{8, 10, 11\}$. This quantization procedure allows us to assess the visual quality and size of our data with regard to different resolution levels, and how it manifests into applications where the data has to be conveyed.
Therefore, we assess our three different types of point clouds, as outlined in ~\cref{sub:colored-pcs}, with a rate-distortion (RD)-based approach in order to evaluate three key outcomes: 1) evaluate how the different characteristics (number of points, density, "watertightness", etc.) from these three types are materialized in a transmission context; 2) observe which voxelized resolutions prove to be more adequate in a live-streaming case according to the data that we generated; 3) assess the conveyance of our data both in an inter-frame and an intra-frame scenario.  

To address objective number 3), 4 consecutive frames of each of our 27 scenes were selected to be conveyed through V-PCC and G-PCC. Further information about the selection of the frames from the scenes and the specifications used for both codecs are explained in the supplementary material.

The naming convention for our results is based on the type of the point cloud, the codec used, and the resolution that was applied to the data, with the format \textit{hset-$codec$-$pc_{type}$-$b$}, where $pc_{type} =$ \{$raw$, $post\_proc$, $sampled$\}, $codec =$ \{$gpcc$, $vpcc$\} and $b = \{8, 10, 11\}$. The RD results are constructed such that the rate consists of the total rate, in bits, required for transmission. For the quality metric, we evaluate the geometry by using the point-to-point metric, also known as D1 \cite{geometric_distortion}, and for the attributes, we use the Peak-Signal-to-Noise-Ratio (PSNR) for the Y, U, and V channels of the original and decoded point clouds.

The results for the average D1 metric from 4 frames of our 27 sequences can be observed in ~\cref{fig:d1-rd-results}, while the results for the attributes are presented and discussed in the supplementary material. Notice that, even though the bitrate increase is expected and increases with the resolution, the conveyed size increases more significantly when going from a resolution of 10 to 11 bits, which is particularly noticeable for the \textit{hset-vpcc-raw} point clouds due to their lower density in comparison to \textit{hset-vpcc-$post\_proc$} and \textit{hset-vpcc-sampled}. The significant decrease in the D1 metric also indicates that the number of decoded points is much larger than the original voxelized point clouds. You can observe this effect for $b$ = 10 for the \textit{hset-vpcc-raw} and $b$ = 11 for \textit{hset-vpcc-$post\_proc$} and \textit{hset-sampled}, with the same happening for G-PCC in a lesser degree. Finally, note that V-PCC provides a better RD performance over G-PCC for the 4 transmitted frames, in particular for $b = 10$. This is to be expected, due to V-PCC's inter-frame scope.


\begin{table*}[!ht]
  \centering
  \caption{Accuracy of FEC models for our datasets (HEADSET-VoCap and HEADSET-LF) compared to available benchmark datasets for FEC.}
  \resizebox{1\textwidth}{!}{%
  \begin{tabular}{l p{2cm} p{2cm} p{1cm} p{1.5cm} p{1.5cm} p{1.5cm}}
    \toprule
    \textbf{Model} & \textbf{HEADSET-VoCap} & \textbf{HEADSET-LF} & \textbf{JAFFE} ~\cite{lyons1998coding} & \textbf{AffectNet} ~\cite{mollahosseini2017affectnet} & \textbf{AFEW} ~\cite{dhall2019emotiw} & \textbf{VGAF} ~\cite{sharma2021audio}\\ \hline
    EfficientNet-B0, 8 classes ~\cite{savchenko2021facial} & 58.19 & 61.98 & 46.00 & 60.10 & 55.14 & 68.29 \\ \hline
    EfficientNet-B2, 7 classes ~\cite{savchenko2022classifying} & 67.44 & 62.25 & 54.00 & 64.30 & 59.63 & 69.84 \\ \hline
    EfficientNet-B2, 8 classes ~\cite{savchenko2022video} & 62.46 & 61.50 & 54.00 & 60.90 & 57.78 & 70.23 \\ \hline
    ResMaskingNet, 7 classes ~\cite{pham2021facial} & 51.11 & 53.34 & 46.95 & 49.81 & -- & -- \\ \hline
    Ad-Corre, 7 classes ~\cite{fard2022ad} & 46.90 & 50.78 & 41.31 & 54.07 & -- & -- \\
    \bottomrule
  \end{tabular}
  }
  \vspace{-2ex}
  \label{tab:result}
\end{table*}

\subsection{Facial expression classification (FEC)}

As described, the MTCNN detector without any margins is utilized before applying the FEC models, so that most parts of the background such as hair follicles are not present. As a result, the learned facial features are more suitable for emotional analysis. 

To make comparisons for evaluation in FEC, we selected four similar datasets namely JAFFE ~\cite{lyons1998coding}, AffectNet ~\cite{mollahosseini2017affectnet}, AFEW ~\cite{dhall2019emotiw}, and VGAF ~\cite{sharma2021audio}, which are used as benchmark by prior works on the FEC problem. In the following, we briefly explain each dataset's characteristics with regard to its content. 

\textit{JAFFE} ~\cite{lyons1998coding}:  The Japanese Female Facial Expression (JAFFE) database contains 213 grayscale images of acted Japanese female facial emotions. All the images are resized into ($256\times256$ pixel). It includes 7 basic human facial emotions (\textit{Anger, Disgust, Fear, Happiness, Sadness, Surprise,} and \textit{Neutral}). For the comparison, we used all 213 images as the testing set.

\textit{AffectNet} ~\cite{mollahosseini2017affectnet}: This RGB image dataset includes 287,651 images in its training set, and 3,999 images in its validation set of 8 human facial expressions (\textit{Anger, Disgust, Fear, Happiness, Sadness, Surprise, Contempt,} and \textit{Neutral}). We used 7 classes (excluding \textit{Contempt} emotion) of the original validation set of 3,499 images for testing purposes of AffectNet. The faces are detected by the authors of the dataset before evaluation. 

\textit{AFEW} ~\cite{dhall2019emotiw}: The AFEW dataset with 773 train and 383 validation samples contains audio-video short clips acquired from TV serials and movies with different poses, spontaneous expressions, and illuminations. They are grouped by a single emotion label to the video clip from 6 basic emotions (\textit{Anger, Disgust, Fear, Happiness, Sadness,} and \textit{Surprise}) and \textit{Neutral}, as described in ~\cite{savchenko2022classifying}.

\textit{VGAF} ~\cite{sharma2021audio}: This dataset shows a wide amount of variations in both training and validation sets. The data contains 2,661 clips and 766 videos for training and validation, respectively. The FEC problem in VGAF is to classify each video into 3 classes (\textit{Positive, Neutral,} and \textit{Negative} emotions)~\cite{savchenko2022classifying}.

Then, we applied five state-of-the-art methods for facial expression classification on the aforementioned and our HEADSET dataset. These methods include the models Ad-Corre (trained on RAF-DB ~\cite{li2017reliable} dataset for 7 classes) ~\cite{fard2022ad}, ResMaskingNet (trained on FER-2013 dataset ~\cite{goodfellow2013challenges} for 7 classes) ~\cite{pham2021facial}, as well as lightweight EfficientNet-B0 (trained for 8 classes), EfficientNet-B2 (trained for 8 classes), and EfficientNet-B2 (trained for 7 classes) ~\cite{savchenko2022classifying,savchenko2022video,savchenko2021facial}, which were trained on the VGGFace2 dataset ~\cite{cao2018vggface2}.
The accuracy performance metric is then computed for all datasets.
~\cref{tab:result} gives a summary of the accuracy measures for all five models on our datasets as well as on the validation sets of AFEW~\cite{dhall2019emotiw}, AffectNet~\cite{mollahosseini2017affectnet}, VGAF~\cite{sharma2021audio}, and all images of the JAFFE dataset~\cite{lyons1998coding}. As EfficientNet-B0 and EfficientNet-B2 are capable of extracting emotional features in video frames ~\cite{savchenko2022classifying}, AFEW and VGAF datasets have only been used for video-based emotion recognition.
The details of F1 scores are presented in the supplementary material due to space limitations.



It is worth mentioning that the usage of a model trained on 8 or 7 classes to predict 7 or 6 emotional categories presents a slightly lower accuracy, though it is more general as it can be used to predict either 8, 7, or 6 emotions ~\cite{savchenko2021facial}.



As proved in ~\cite{chiesa2018multi}, multi-view representation of light field images recorded by a Lytro Illum camera can provide complementary information beneficial for face recognition. Thus, we have processed the raw data of Lytro Illum camera to render LF images as multi-view RGB images, each one collected from a slightly shifted point of view. For this experiment, each data is transformed in a 5x5 RGB view matrix, each view with size $620 \times 432$ pixels.
As it is observable in ~\cref{tab:result}, the results of all five models on our datasets are comparable to other benchmark datasets in FEC.

\subsection{HMD removal}

For facial reconstruction of the areas occluded by an HMD in the collected video frames, we used a GAN-based method for deep video inpainting, named Learnable Gated Temporal Shift Module (LGTSM) introduced in \mbox{~\cite{chang2019learnable}}, with the same hyperparameters and training procedure. The following modifications were applied. We added an additional self-attention layer and its non-local operations, introduced in \mbox{~\cite{zhang2019self}}, in its encoder part before the dilated convolution layers in order to capture long-range dependencies between different regions of an input feature map. Specifically, the self-attention module takes a feature map as input and applies three convolution layers to compute the key, query, and value vectors. The kernel size and stride for each convolutional layer are set to $1\times1$ and 1, respectively, to capture spatial relationships between neighboring feature map locations while keeping computational costs  low.

We used FaceForensics dataset ~\cite{rossler2018faceforensics} and our collected RGB data from VoCap as the training set and validation set, respectively. FaceForensics is comprised of 1,004 videos including more than 500,000 frames with faces collected from Youtube. They consist of only frontal faces cropped to $128\times128$. 
The whole RGB video sequences in our dataset collected from VoCap include 11,584 frames captured by camera number 30 and 1 from all 27 volunteers, as described in ~\cref{subsec:RGBD}. We first applied MTCNN for face recognition, then resized the frames to $128\times128$ to make them similar to the training set.

\begin{figure}[!ht]
\centering
\includegraphics[width=0.9\linewidth]{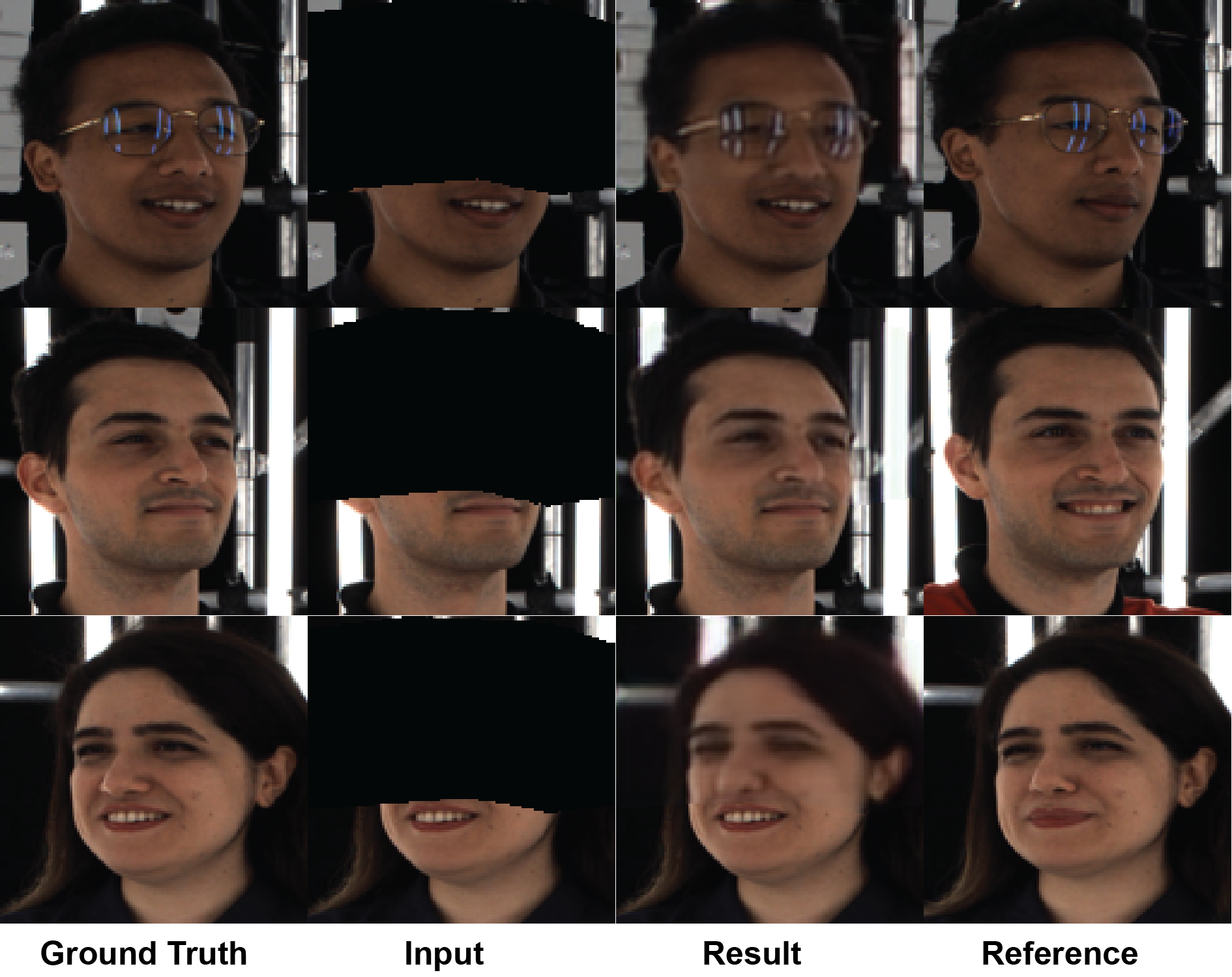}
\caption{Face completion of three different participants from HEADSET. The images are ground truth, input frame, inpainted result, and occlusion-free reference image, respectively.}
\label{fig:HMDRem}
\end{figure}

For preparing HMD masks on the video frames, we first created binary masks of the VR headset captured in experiment C. Then, we applied the masks to the ground truth images to be the input data for the inpainting network.
We also used the first frames from each video sequence without any occlusion as a reference image that imposes an identity prior to the searching space of the network. The reference images were fed into the network jointly with the masked frames.
      



Samples of the qualitative results of the LGTSM model with a self-attention module on HEADSET are illustrated in ~\cref{fig:HMDRem}. The examples are from three different individuals and captured by two distinct cameras. The first two rows are the inpainted results from starting frames, and the last row demonstrates an illustration of HMD removal outputs from the final frames. 
While the qualitative findings are promising, they exhibit temporal inconsistencies across frames, underscoring the need for additional research to comprehensively investigate potential strategies for mitigating HMD occlusion removal.
\section{Conclusion}
\label{sec:conclusion}
In this work, we have captured and presented a multimodal database that depicts humans performing posed and spontaneous facial expressions and subtle body movements. We have also recorded a part of the database with HMD occlusion. Our capturing setup includes a VoCap studio and a Lytro Illum camera. On top of the obtained textured meshes, colored point clouds, multi-view RGB-D images, and light field images, the raw captured data, calibration, and camera parameters are also made available. The proposed databases’ performance in comparison with similar datasets has also been evaluated in different application scenarios. The provided material can facilitate the design of immersive media technologies and XR applications in which realistic human interaction is necessary. We believe that our database will then promote further research in data-driven techniques, computer vision for XR, human interactions in XR, and volumetric data reconstruction by providing a high-quality testing set for performance evaluation. Although the utilization of HEADSET holds significant potential for advancing research pertaining to XR applications, an extension of the existing HEADSET version can further enhance the progress of these technologies. This extension encompasses the inclusion of a greater number of participants, extended recording capabilities, as well as an examination of the impact of diverse factors such as age and medical conditions on individuals' perceived emotions.


\acknowledgments{%
This project has received funding from the European Union’s Horizon 2020 research and innovation program under the Marie Skłodowska-Curie grant agreement No 956770. The data collection part was carried out with the support of Centre for Immersive Visual Technologies (CIVIT) research infrastructure, Tampere University, Finland. We want to especially thank Jani Käpylä, for his help during the capturing. 
}

\bibliographystyle{abbrv-doi-hyperref}

\bibliography{refs}


\end{document}